\documentclass{bmvc2k}

\usepackage{times}
\usepackage{epsfig}
\usepackage{graphicx}
\usepackage{amsmath}
\usepackage{amssymb}
\usepackage{booktabs}
\usepackage{multirow}
\usepackage{rotating}
\usepackage{tabularx}
\usepackage{xcolor}
\usepackage{stmaryrd}
\usepackage{pifont}%
\usepackage{setspace}
\usepackage{makecell}
\newcommand{\xmark}{\ding{55}}%

\title{COARSE3D: Class-Prototypes for Contrastive Learning in~Weakly-Supervised 3D~Point~Cloud Segmentation}

\addauthor{Rong Li}{selirong@mail.scut.edu.cn}{1}%
\addauthor{Anh-Quan Cao}{anh-quan.cao@inria.fr}{2}
\addauthor{Raoul de Charette}{raoul.de-charette@inria.fr}{2}

\addinstitution{
 South China University of Technology\\
 Guangzhou, China
}
\addinstitution{
 Inria\\
 Paris, France
}
\runninghead{Li, Cao, de Charette}{COARSE3D}

\makeatletter
\newcommand*\@dblLabelI {}
\newcommand*\@dblLabelII {}
\newcommand*\@dblequationAux {}

\def\@dblequationAux #1,#2,%
{\def\@dblLabelI{\label{#1}}\def\@dblLabelII{\label{#2}}}

\newcommand*{\doubleequation}[3][]{%
	\par\vskip\abovedisplayskip\noindent
	\if\relax\detokenize{#1}\relax
	\let\@dblLabelI\@empty
	\let\@dblLabelII\@empty
	\else %
	\@dblequationAux #1,%
	\fi
	\makebox[0.5\linewidth-1.5em]{%
		\hspace{\stretch2}%
		\makebox[0pt]{$\displaystyle #2$}%
		\hspace{\stretch1}%
	}%
	\makebox[0.5\linewidth-1.5em]{%
		\hspace{\stretch1}%
		\makebox[0pt]{$\displaystyle #3$}%
		\hspace{\stretch2}%
	}%
	\makebox[3em][r]{(%
		\refstepcounter{equation}\theequation\@dblLabelI, 
		\refstepcounter{equation}\theequation\@dblLabelII)}%
	\par\vskip\belowdisplayskip
}
\makeatother

\newcolumntype{P}[1]{>{\centering\arraybackslash}p{#1}}

\newcommand{\tablesize}[0]{\scriptsize}

\DeclareFontFamily{U}{mathb}{}
\DeclareFontShape{U}{mathb}{m}{n}{
  <-5.5> mathb5
  <5.5-6.5> mathb6
  <6.5-7.5> mathb7
  <7.5-8.5> mathb8
  <8.5-9.5> mathb9
  <9.5-11.5> mathb10
  <11.5-> mathb12
}{}
\DeclareSymbolFont{mathb}{U}{mathb}{m}{n}
\DeclareMathSymbol{\ulsh}{3}{mathb}{"E8}
\DeclareMathSymbol{\ursh}{3}{mathb}{"E9}
\DeclareMathSymbol{\dlsh}{3}{mathb}{"EA}
\DeclareMathSymbol{\drsh}{3}{mathb}{"EB}

\def\eg{\emph{e.g}\bmvaOneDot}

\newcommand{\add}[1]{\textcolor{orange}{#1}}
\newcommand{\edit}[2]{\textcolor{orange}{#2}}

\newcommand{\ourmethod}[1]{\textcolor{red}{OurFancyMethod}}

\definecolor{car}{rgb}{0.39215686, 0.58823529, 0.96078431}
\definecolor{bicycle}{rgb}{0.39215686, 0.90196078, 0.96078431}
\definecolor{motorcycle}{rgb}{0.11764706, 0.23529412, 0.58823529}
\definecolor{truck}{rgb}{0.31372549, 0.11764706, 0.70588235}
\definecolor{other-vehicle}{rgb}{0.39215686, 0.31372549, 0.98039216}
\definecolor{person}{rgb}{1.        , 0.11764706, 0.11764706}
\definecolor{bicyclist}{rgb}{1.        , 0.15686275, 0.78431373}
\definecolor{motorcyclist}{rgb}{0.58823529, 0.11764706, 0.35294118}
\definecolor{road}{rgb}{1.        , 0.        , 1.        }
\definecolor{parking}{rgb}{1.        , 0.58823529, 1.        }
\definecolor{sidewalk}{rgb}{0.29411765, 0.        , 0.29411765}
\definecolor{other-ground}{rgb}{0.68627451, 0.        , 0.29411765}
\definecolor{building}{rgb}{1.        , 0.78431373, 0.        }
\definecolor{fence}{rgb}{1.        , 0.47058824, 0.19607843}
\definecolor{vegetation}{rgb}{0.        , 0.68627451, 0.        }
\definecolor{trunk}{rgb}{0.52941176, 0.23529412, 0.        }
\definecolor{terrain}{rgb}{0.58823529, 0.94117647, 0.31372549}
\definecolor{pole}{rgb}{1.        , 0.94117647, 0.58823529}
\definecolor{traffic-sign}{rgb}{1.        , 0.        , 0.    }

\definecolor{nus00}{rgb}{0.9607843137254902, 0.9411764705882353, 1.0}
\definecolor{nus01}{rgb}{0.5647058823529412, 0.5019607843137255, 0.4392156862745098}
\definecolor{nus02}{rgb}{0.23529411764705882, 0.0784313725490196, 0.8627450980392157}
\definecolor{nus03}{rgb}{0.0, 0.27058823529411763, 1.0}
\definecolor{nus04}{rgb}{0.0, 0.6196078431372549, 1.0}
\definecolor{nus05}{rgb}{0.27450980392156865, 0.5882352941176471, 0.9137254901960784}
\definecolor{nus06}{rgb}{0.38823529411764707, 0.23921568627450981, 1.0}
\definecolor{nus07}{rgb}{0.5019607843137255, 0.0, 0.0}
\definecolor{nus08}{rgb}{0.30980392156862746, 0.30980392156862746, 0.1843137254901961}
\definecolor{nus09}{rgb}{0.0, 0.5490196078431373, 1.0}
\definecolor{nus10}{rgb}{0.2784313725490196, 0.38823529411764707, 1.0}
\definecolor{nus11}{rgb}{0.7490196078431373, 0.8117647058823529, 0.0}
\definecolor{nus12}{rgb}{0.29411764705882354, 0.0, 0.6862745098039216}
\definecolor{nus13}{rgb}{0.29411764705882354, 0.0, 0.29411764705882354}
\definecolor{nus14}{rgb}{0.23529411764705882, 0.7058823529411765, 0.4392156862745098}
\definecolor{nus15}{rgb}{0.5294117647058824, 0.7215686274509804, 0.8705882352941177}
\definecolor{nus16}{rgb}{0.0, 0.6862745098039216, 0.0}

\definecolor{poss01}{rgb}{ 0.11764705882352941, 0.11764705882352941, 1.0 }
\definecolor{poss02}{rgb}{ 0.7843137254901961, 0.1568627450980392, 1.0 }
\definecolor{poss03}{rgb}{ 0.9607843137254902, 0.5882352941176471, 0.39215686274509803 }
\definecolor{poss04}{rgb}{ 0.0, 0.23529411764705882, 0.5294117647058824 }
\definecolor{poss05}{rgb}{ 0.0, 0.6862745098039216, 0.0 }
\definecolor{poss06}{rgb}{ 0.0, 0.0, 1.0 }
\definecolor{poss07}{rgb}{ 0.5882352941176471, 0.9411764705882353, 1.0 }
\definecolor{poss08}{rgb}{ 0.0, 1.0, 0.49019607843137253 }
\definecolor{poss09}{rgb}{ 0.0, 0.7843137254901961, 1.0 }
\definecolor{poss10}{rgb}{ 1.0, 1.0, 0.19607843137254902 }
\definecolor{poss11}{rgb}{ 0.19607843137254902, 0.47058823529411764, 1.0 }
\definecolor{poss12}{rgb}{ 0.9607843137254902, 0.9019607843137255, 0.39215686274509803 }
\definecolor{poss13}{rgb}{ 0.5019607843137255, 0.5019607843137255, 0.5019607843137255 }

\makeatletter
\newcommand{\car@semkitfreq}{4.08}
\newcommand{\bicycle@semkitfreq}{0.02}
\newcommand{\motorcycle@semkitfreq}{0.04}
\newcommand{\truck@semkitfreq}{0.21}
\newcommand{\othervehicle@semkitfreq}{0.16}
\newcommand{\person@semkitfreq}{0.18}
\newcommand{\bicyclist@semkitfreq}{1.11e-6}
\newcommand{\motorcyclist@semkitfreq}{5.53e-9}
\newcommand{\road@semkitfreq}{19.87}  %
\newcommand{\parking@semkitfreq}{1.47}
\newcommand{\sidewalk@semkitfreq}{14.39}  %
\newcommand{\otherground@semkitfreq}{0.39}
\newcommand{\building@semkitfreq}{13.26}  %
\newcommand{\fence@semkitfreq}{7.23}
\newcommand{\vegetation@semkitfreq}{26.68}  %
\newcommand{\trunk@semkitfreq}{0.60}
\newcommand{\terrain@semkitfreq}{7.81} %
\newcommand{\pole@semkitfreq}{0.28}
\newcommand{\trafficsign@semkitfreq}{0.06}
\newcommand{\semkitfreq}[1]{{\csname #1@semkitfreq\endcsname}}

\newcommand{\paragraphcond}[1]{\vspace{0.3em}\noindent\textbf{#1}}

\begin{document}

\maketitle
\vspace{-0.5em}
\begin{abstract}
	Annotation of large-scale 3D data is notoriously cumbersome and costly. As an alternative, weakly-supervised learning alleviates such a need by reducing the annotation by several order of magnitudes.
	We propose COARSE3D, a novel architecture-agnostic contrastive learning strategy for 3D segmentation.
	Since contrastive learning requires rich and diverse examples as \emph{keys} and \emph{anchors}, we leverage a \emph{prototype memory bank} capturing class-wise global dataset information efficiently into a small number of prototypes acting as \emph{keys}. An \emph{entropy-driven sampling} technique then allows us to select good pixels from predictions as \emph{anchors}.
	Experiments on three projection-based backbones show we outperform baselines on three challenging real-world outdoor datasets, working with as low as 0.001\% annotations.

\end{abstract}

\section{Introduction}\vspace{-0.2em}
\label{sec:intro}

Semantic segmentation is the holy task of any scene understanding system. For driving, this is of utmost importance, because driving requires a fine-grained 3D understanding for navigation and planning. Subsequently, in recent years, 3D semantic segmentation attracted a plethora of works~\cite{cortinhal2020salsanext, randlanet, squeezesegv1, cylinder3d}.
However, the vast majority of the literature relies on supervised learning -- thus assuming dense point-wise semantic labels -- which are laborious and costly to acquire. A striking example is the popular urban dataset SemKITTI~\cite{semantickitti}, which took more than 1700 hours to label by human operators.

To alleviate this cost, weakly-supervised 3D segmentation uses several orders of magnitude fewer labels \cite{sqn, liu2021one, redal,wei2020multi}.
Still, researches are mostly focused on point-based backbones which are efficient but often slow at inference.
Conversely, projection-based methods~\cite{rangenet++, aksoy2020salsanet, cortinhal2020salsanext, squeezesegv3} were shown to operate at high speed with competitive performance~\cite{cortinhal2020salsanext}.

In this work, we propose an architecture-agnostic contrastive learning framework for weakly-supervised semantic segmentation.
Since our approach does not alter the backbone segmentation, we demonstrate its usage on a lightweight projection backbone, thus enabling both efficient trainings with extreme scarce labels and fast inference.
Of paramount importance, contrastive learning needs abundant representative examples~\cite{he2020momentum, chen2020improved, wu2018unsupervised} which is non-trivial to get in weakly-supervised settings. To solve this we introduce two mechanisms.
First, we propose a \emph{prototype memory bank} that stores the global dataset per-class information into a small number of rich and compact prototypes.
Second, we employ an entropy-driven sampling to select sufficiently good \emph{anchors} from inaccurate predictions.
Experiments show that our method outperform existing weakly-supervised baselines on three real-world outdoor datasets SemKITTI~\cite{semantickitti}, SemPOSS~\cite{pan2020semanticposs}, and nuScenes~\cite{nuScenes}. On SemKITTI we also show that we \textit{on par} with full-supervision (i.e., 100\% labels) performance, using only 1\% of the labels.
\textbf{Our code is released at: \url{https://github.com/cv-rits/COARSE3D}}.
\\[0.5em]
\noindent{}In brief, our contributions can be summarized as follows:
\vspace{-0.4em}\begin{itemize}
	\setlength{\itemsep}{1pt}
	\setlength{\parskip}{0pt}
	\setlength{\parsep}{0pt}
	\item We propose an architecture-agnostic framework for weakly-supervised 3D point cloud segmentation and, to the best of our knowledge, first, demonstrate its usage on a lightweight projection backbone.
	\item We introduce a \emph{prototype memory bank} that captures per-class dataset information with an \textit{entropy-driven sampling} technique to sample more confident pixels as anchors.
	\item Our strategy significantly outperforms all comparable baselines on three real-world outdoor datasets, with as few labels as 0.001\% -- roughly corresponding to a single labelled point per frame.
\end{itemize}

\vspace{-0.5em}
\section{Related Work}\vspace{-0.2em}

\noindent\textbf{Label-efficient 3D Semantic Segmentation.}
With the abundant annotated large-scale 3D datasets~\cite{semantickitti, pan2020semanticposs, nuScenes, Liao2021ARXIV, dai2017scannet}, point cloud semantic segmentation advanced rapidly. The methods can be categorized into point-based~\cite{2s3net,spvnas, latticenet, randlanet, tangentconv, splatnet, spg} and projection-based~\cite{cortinhal2020salsanext, aksoy2020salsanet, squeezesegv3, squeezesegv2, squeezesegv1, rangenet++} approaches.
To learn from fewer labels, various strategies were proposed with point-based methods: enforcing geometric prior~\cite{xu2020weakly}, local neighborhood propagation~\cite{sqn, liu2021one}, for smarter use of the rare labels some used active learning ~\cite{redal}, pseudo labelling~\cite{wei2020multi, li2022hybridcr, shi2022weakly}, graph propagation~\cite{liu2021one, shi2022weakly}, self-training~\cite{li2022hybridcr, zhang2021perturbed}, temporal matching~\cite{shi2022weakly, unal2022scribble}. \edit{Still, there is no approaches designed for the 2D nature of the projection-based methods.}{}

\noindent\textbf{Contrastive Learning for 3D.}
Contrastive learning~\cite{SimCLR, chen2020improved, he2020momentum}, initially designed for 2D, was extended to 3D using similar points from different views in~\cite{zhang2021self, xie2020pointcontrast, hou2021exploring, lal2021coconets}.
Limited training data is considered in~\cite{hou2021exploring}.
Invariant representation is learned from spatio-temporal cues in~\cite{chen20214dcontrast, huang2021spatio, liang2021exploring}.
Pairs of point and pixels regions were matched in~\cite{Sautier_2022_CVPR} while point-pixel pairs are regarded as a whole in~\cite{liu2020p4contrast}.
We focus the literature on crucial aspects for our work.

\noindent\textit{Sampling Strategy} is crucial for constrastive learning that requires many good examples~\cite{kaya2019deep, chen2020improved, he2020momentum, wu2018unsupervised}. Prior knowledge is used to guide sampling in~\cite{hadsell2006dimensionality, bell2015learning}.
Multiple pairs of positive and negative data are sampled in~\cite{bucher2016improving, sohn2016improved}. To improve convergence, hard negative sampling is exploited in~\cite{simo2015discriminative, bucher2016hard, schroff2015facenet, khosla2020supervised, robinson2020contrastive, kalantidis2020hard} and point proxies serve as examples in~\cite{movshovitz2017no}. To avoid local minima, semi-hard negative sampling is proposed in~\cite{schroff2015facenet, cai2020all, xie2020delving}. In~\cite{cui2016fine}, false positive images are used as hard negative examples. Since noisy samples were shown to deteriorate performance~\cite{wu2017sampling}, hard positive/negative examples sampling is applied in \cite{khosla2020supervised} while high-quality anchors are sampled by considering correct prediction \add{in}~\cite{cross_img_contrast}.
We instead propose entropy-driven anchors sampling techniques to learn more robust representations.

\noindent{}\textit{Memory Bank} introduced in~\cite{wu2018unsupervised} allows storing many examples, \eg pixels in 2D, typically sampled per image~\cite{xie2021propagate, wang2021dense, chaitanya2020contrastive} or mini-batch~\cite{cross_img_contrast} to increase variety.
However, such storage comes at the cost of large memory and asynchronous updates.
These issues are mitigated using momentum encoder~\cite{chen2020improved, he2020momentum} or keeping examples in recent batches~\cite{chen2020improved, he2020momentum, wang2020cross}.
Pixels are selected randomly~\cite{cross_img_contrast} or selectively~\cite{alonso2021semi} from each image.
Still, only a few pixels
are stored due to memory constraints, losing full dataset content. Moreover, using pixels embeddings
is redundant~\cite{cross_img_contrast}. Regions~\cite{cross_img_contrast, hu2021region} can be employed as more compact representation but not with label sparsity settings due to the need of per-pixel annotation.
Hence, we propose an efficient memory bank, storing prototypes (instead of pixels) that can capture the class-wise dataset-wise information using only a small number of prototypes.

\noindent\textit{Prototype Learning} aims to represent the embedding space as a set of prototypes. It is used for many tasks like classification~\cite{cover1967nearest, garcia2012prototype, goldberger2004neighbourhood, salakhutdinov2007learning, guerriero2018deepncm, mettes2019hyperspherical, Wu2018ECCV, yang2018robust}, few-/zero-shot~\cite{snell2017proto,jetley2015prototypical, bucher2016hard}, regression~\cite{mettes2019hyperspherical}, unsupervised learning~\cite{wu2018unsupervised, xu2020attribute}, explanability~\cite{li2018deep}, and semantic segmentation~\cite{zhou2022rethinking, dong2018few, vinyals2016matching, wang2019panet}.
Different from the literature, we use prototypes as a memory bank to gather the rich class-wise global context information from the whole dataset efficiently. \\[-1.5em]

\vspace{-0.5em}
\section{Proposed approached}\vspace{-0.2em}
We address the problem of 3D semantic segmentation of point clouds and propose, a novel architecture-agnostic training strategy for weakly-supervised learning coined `COARSE3D'.
To learn efficient semantic representation from scarce labels, we leverage a custom contrastive learning module whose originality resides in a lightweight memory footprint and sparse embedding optimization.
Results show our methodology is efficient with as little as $0.001\%$ of the labelled data
-- stepping ahead of prior works~\cite{sqn, xu2020weakly, zhang2021weakly, zhang2021self}.

\begin{figure*}
	\begin{center}
		\includegraphics[width=0.95\linewidth]{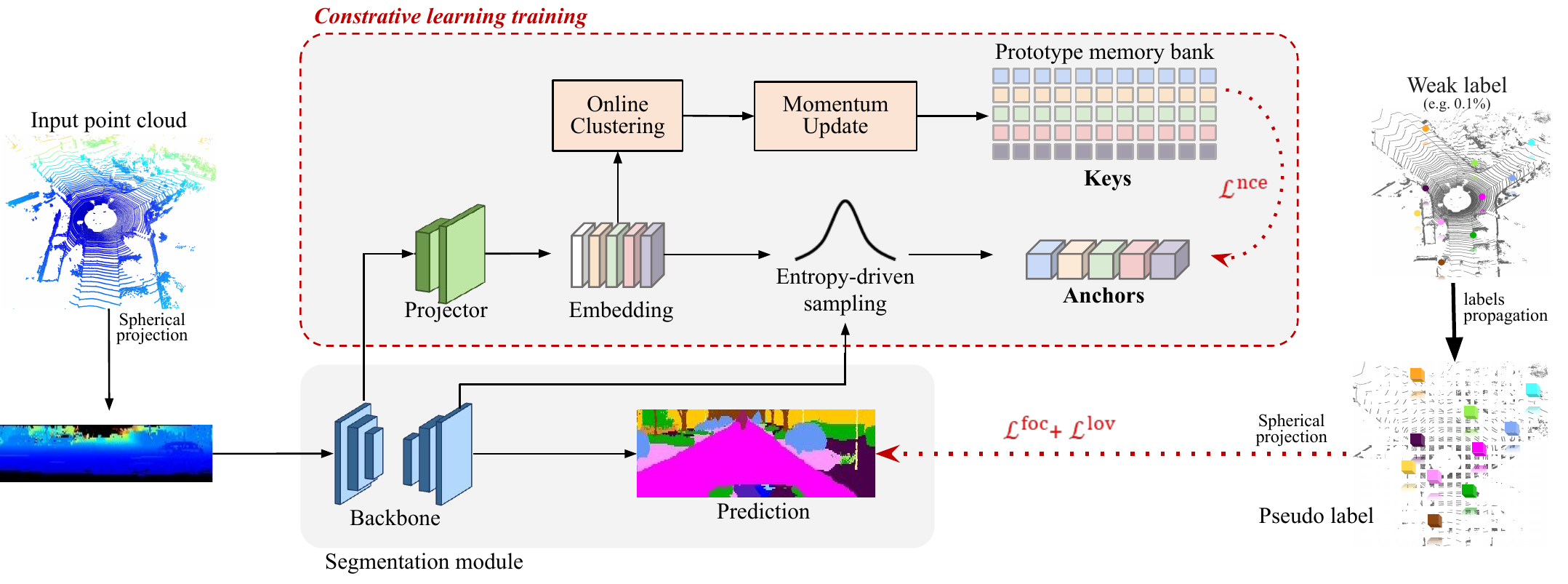}

	\end{center}
	\vskip -1.5em
	\caption{
		Our COARSE3D uses a contrastive learning strategy (Sec.~\ref{sec:pix2proto_contrastive_learning}) to learn high dimensional semantic embeddings from the encoder features, later clustered into a lightweight prototype memory bank (Sec.~\ref{sec:proto_mem_bank}).
		At each step, contrastive learning is applied on a scarce set of `anchors', selected from an entropy-based sampling (Sec.~\ref{sec:achors_sampling}), which features are pushed towards (pulled away from) positive (negative) prototypes acting as `keys'.
	}
	\label{fig:arch}
	\vspace{-1em}
\end{figure*}

COARSE3D, depicted in Fig.~\ref{fig:arch}, uses a custom contrastive module (Sec.~\ref{sec:pix2proto_contrastive_learning}) that projects features of the segmenter encoder into a different embedding space.
As for the segmenter backbone, we employ the fast and memory efficient SalsaNext~\cite{cortinhal2020salsanext} which processes 3D point cloud as 2D range images.
Rather than the greedy pixel-wise memory~\cite{cross_img_contrast}, our projected features are clustered into a prototype memory bank (Sec.~\ref{sec:proto_mem_bank}) that captures compact class-wise semantic embeddings. This enables preserving the dataset context in a significantly lighter fashion.
In each iteration, we optimize only a small subset of pixels (i.e. anchors) sampled with an entropy-based strategy (Sec.~\ref{sec:achors_sampling}) considering prototypes as positive/negative samples (i.e. keys).
Our training strategy (Sec.~\ref{sec:meth_training_strategy}) combines standard segmentation and contrastive losses, along with a simple labels propagation strategy.\vspace{-0.5em}

\subsection{Contrastive Learning Pipeline}\vspace{-0.2em}
\label{sec:pix2proto_contrastive_learning}

We leverage a contrastive learning scheme since prior works have shown its effectiveness to learn from limited labels~\cite{xie2020pointcontrast, rozenberszki2022language, hou2021exploring}.
We build on the standard InfoNCE pixel-wise contrastive loss~\cite{CPCinfoNCE, gutmann2010noisecontrast} but apply the latter between our class-wise representations, hereafter \textit{prototypes} (Sec.~\ref{sec:proto_mem_bank}), and a subset of sampled pixels (Sec.~\ref{sec:achors_sampling}).

More specifically, at each training iteration we select a set of pixels, called \textit{anchors} having features $\mathcal{A}{=}\{a_j,..., a_{N_\text{a}}\}$. For each anchor, we select the corresponding \emph{prototypes} of similar semantic class, i.e. \textit{positive keys} with features $\mathcal{P}^{+}{=}\{p^{+}_1, ..., p^{+}_{N^{+}}\}$, or different semantic class, i.e. \textit{negative keys} with features $\mathcal{P}^{-}{=}\{p^{-}_1, ..., p^{-}_{N^{-}}\}$.
The InfoNCE optimization objective is to simultaneously pull the embedding of \textit{anchors} towards \textit{positive keys} while pushing them away from the \textit{negative keys}. Formally, it writes:
\begin{equation}
	\mathcal{L}^{\text{Pix2Proto}}=  \frac{1}{N_\text{a}} \sum_{a_i \in \mathcal{A} } -\text{log} \frac{\sum_{p_j^+ \in \mathcal{P}^{+}}\exp \left(a_i \cdot p_j^{+} / \tau\right)}{\sum_{p_j^{+}  \in \mathcal{P}^{+} }\exp \left(a_i \cdot p_j^{+} / \tau\right)+\sum_{p_j^{-}  \in \mathcal{P}^{-} } \exp \left(a_i \cdot p_j^{-} / \tau\right)}\,,
	\label{eq:pixel_proto}
\end{equation}
where $\tau$ is the temperature hyperparameter. All feature vectors are extracted by first linearly interpolating multi-scale features to the full image size. Then, they are concatenated and mapped to 256-dim using two $1{\times}1$ 2D convolutional layers. Finally, they are $\ell^2$-normalized.
Following previous works~\cite{kaya2019deep, cross_img_contrast} that demonstrated the benefit of a custom selection of keys and anchors, we consider a set of anchors sampled using the entropy of the segmentation output and keys from a newly introduced prototype memory bank, that we describe next.

\vspace{-0.5em}
\subsection{Prototype Memory Bank}\vspace{-0.2em}
\label{sec:proto_mem_bank}

The memory bank requires massive data to learn good representation which is traditionally achieved by storing a large number of pixels as in~\cite{he2020momentum, cross_img_contrast, wu2018unsupervised, wang2020cross}. Still, this is very costly in terms of memory and computation for contrastive sampling. Furthermore, storing pixels is semantically redundant which can be mitigated by storing semantic regions~\cite{cross_img_contrast}. However, the latter would be poorly reliable given the limited labels.
Therefore, we introduce a \emph{prototype memory bank} that encodes class-wise embedding into compact prototypes.

More in-depth, the memory bank stores $\{P_k \in \mathbb{R}^{N_p{\times}D}\}^K_{k=1}$ prototypes of feature dimension $D$, corresponding to $N_p$ prototypes per class for each of the $K$ semantic classes.
For each semantic class $k$, at each training iteration, we cluster the $N_{k}$ labelled pixels of such class into the corresponding $N_p$ class prototypes.
We frame this as an optimal transport problem and employ the Sinkhorn algorithm~\cite{cuturi2013sinkhorn}, similar to~\cite{zhou2022rethinking} because it balances the pixels assignment between all prototypes. Here, note that the prototypes are solely updated from the embeddings of labelled pixels which guarantees reliable information.

In pratice, we compute the optimal transport matrix $T \in \mathbb{R}^{N_{k}{\times}N_p}$ by unrolling three iterations of the Sinkhorn algorithm on the cost matrix $\mathcal{C} \in \mathbb{R}^{N_{k}{\times}N_p}$ -- where $\mathcal{C}_{ij}$ is the cosine distance between the feature maps of pixel $i$ and prototype $j$.
The pixel-prototype mapping of pixel $p_i$ is obtained with $m(p_i){=} \text{argmax}_j T_{ij}$.
In practice, to encourage exploration we use a Gumbel-Softmax ($\tau{=}0.5$) as differentiable argmax.
Given a random initialization, we update each prototype with a $\sigma=0.999$ momentum towards the mean of the clustered pixels embeddings in each training iteration.
Hence, \edit{the }{}$j^{th}$ prototype $\{P_k\}_j$ of class $k$ is updated as:
\begin{equation}
	\{P_k\}_j=\sigma \{P_k\}_j +(1-\sigma)\frac{1}{\sum_{i=1}^{N_{k}} \llbracket m(x_i){=}j\rrbracket }
	\sum_{i=1}^{N_{k}}{x_i}\llbracket m(x_i){=}j\rrbracket\,,
	\label{eq:prototype-update}
\end{equation} with $\llbracket.\rrbracket$ the Iverson brackets.

\vspace{-0.5em}
\subsection{Entropy-driven Anchors Sampling}\vspace{-0.2em}
\label{sec:achors_sampling}

Prior works have shown the importance of proper sampling in contrastive learning~\cite{bucher2016hard, kalantidis2020hard, sohn2016improved, cai2020all, cross_img_contrast, xie2020delving, schroff2015facenet}.
However, such strategies are inappropriate in our setup given the limited labels and the need for abundant samples in contrastive learning.
Instead, we build on information theory~\cite{shannon2001mathematical} and introduce an \emph{entropy-driven sampling} technique to select the most appropriate pixels to serve as \textit{anchors} among the abundant pseudo-labels predictions.

Given an image $x$ processed by the segmentation module, we estimate the quality of the semantic prediction for pixel $x_i$ using its Shannon entropy $H(x_i) = - $ $ \sum_{k=1}^{K}f(x_i) \log(f(x_i))$, where $K$ is the number of classes and $f(x_i) \in \mathcal{R}^{K}$ is the softmax prediction.
Subsequently, we define the sampling probability as $\rho(x_i) = \frac{\exp{-{H(x_i)}^2}}{\sum_{x_j \in \mathcal{X}}\exp{-{H(x_j)}^2}}$ where $\mathcal{X}$ is set of pixels with the same predicted label as $x_i$.
Intuitively, pixels with low entropy will have higher chances to be selected. The power of $2$ is used to make the distribution steeper.
Finally, we perform weighted sampling to select the $N_a$ pixels as a set of anchors $\mathcal{A}$.
Unlike standard pixel-wise contrastive loss~\cite{cross_img_contrast} requiring thousands of pixels, we start with only one anchor and linearly increase during training -- up to 50\% pseudo label. This follows the intuition that the segmentation module improves as training progresses.

\vspace{-0.5em}
\subsection{Training Strategy}\vspace{-0.2em}
\label{sec:meth_training_strategy}
To train the method, we first apply a simple voxel propagation scheme to expand labels illustrated in Fig.~\ref{fig:arch} and detailed in Appendix~\ref{sec:implementation-voxel}. 
We then train with two supervised and one unsupervised loss.
First, the focal loss~\cite{focal-loss} is applied on all labelled points to alleviate classes imbalance:
\begin{equation}
	\mathcal{L}_{i}^{\mathrm{foc}}=-\alpha (1-(f(x_i))^{\gamma} \log (f(x_i))\,,
	\label{eq:focloss}
\end{equation}
where $\alpha=\{w_k \mid y_i=k \}$ is the weight of class $k$, $y_i$ the pseudo label of point $i$, and $w_k= \log(1+ \frac{1}{\text{freq}_k})$ with $\text{freq}_k$ is the class frequency before voxel propagation.
Second, we use the lov{\'a}sz-softmax loss~\cite{lov-loss} to directly optimize the mean intersection-over-union, defined as:
\begin{equation}
	\mathcal{L}_{i}^{\mathrm{lov}}=\frac{1}{k} \sum_{k=1}^{K} \overline{\Delta_{J_{k}}}(\tilde{\mathbf{m}}(c))\;\text{, with}\;\;\;\;\tilde{\mathbf{m}}_{i}(k)= \begin{cases}1-f(x_i) & \text { if } k=y_i, \\ f(x_i) & \text { otherwise, }\end{cases}\,
	\label{eq:lovloss}
\end{equation}
where $\overline{\Delta_{J_{s}}}$ indicates the Lov{\'a}sz extension of the Jaccard index for class $k$. $y_i$ is the label of point $x_i$.
Finally, the overall training target is
\begin{equation}
	\mathcal{L}=\sum_{i}\left(
	\lambda_\text{ce} \mathcal{L}_{i}^{\mathrm{foc}}
	+
	\lambda_\text{lov} \mathcal{L}_{i}^{\mathrm{lov}}
	+
	\lambda_\text{nce} \mathcal{L}_{i}^{\mathrm{nce}}
	\right)\,.
\end{equation}

\vspace{-0.5em}
\section{Experiments}
\vspace{-0.2em}
We evaluate our proposal by comparing against recent baselines on 3 challenging real-world LiDAR datasets, being SemKITTI~\cite{semantickitti}, nuScenes~\cite{nuScenes} and SemPOSS\cite{pan2020semanticposs} and report main results in Sec.~\ref{sec:exp-perf} and ablations in Sec.~\ref{sec:exp-ablation-study}.
For memory-efficient processing, we evaluate solely on projection-based backbones but COARSE3D is applicable to any architecture (\textit{e.g.}, point-based). In particular main results use SalsaNext~\cite{cortinhal2020salsanext} backbone and we experiment with Rangenet-21~\cite{rangenet++} and SqueezeSegV3-21~\cite{squeezesegv3} backbones in ablations.
Importantly, all our components are removed at inference, leaving the segmenter backbone unchanged.

\vspace{-0.5em}
\subsection{Experimental Setup}\vspace{-0.2em}
\label{sec:exp-setup}
\paragraphcond{Datasets.}
\textbf{SemanticKITTI}~\cite{semantickitti}~has 22 German scenes recorded at 10Hz with an HDL-64E LiDAR. We follow common practice and train on sequences 00 to 10, except 08 for validation. The \textit{hidden} test set uses sequences 11 to 21.
\textbf{nuScenes}~\cite{nuScenes}~has 850 scenes from U.S.A./Singapore acquired with a VLP-32 LiDAR, labelled at 2Hz. Again, we follow standards using 700 scenes for training and 150 scenes for validation.
\textbf{SemanticPOSS}~\cite{pan2020semanticposs} recorded Chinese campus scenes with @10Hz 40-layers LiDAR. Though much smaller, the dataset has many pedestrians. We train on sequences 00 to 05, except 02 for validation.

\paragraphcond{Model Selection.}
For SemKITTI~\cite{semantickitti}, we cherrypick the best model after hyperparameter tuning on the validation set. The main evaluation of the chosen model is done using an online official benchmark (hidden test set). Evaluations on SemPOSS~\cite{pan2020semanticposs} and nuScenes~\cite{nuScenes} use the exact same setting but \textit{without any} hyperparameters tuning thus reporting only results of the \textit{last training model} on their validation sets.

\paragraphcond{Annotation Strategy.}
To avoid the human labeling efforts of active learning~\cite{redal} or one-point-per-object annotation~\cite{liu2021one}, we apply random subsampling to existing dense ground truths (i.e. 100\% annotation) to obtain sparse labels (\eg 0.01\%) as in~\cite{sqn, xu2020weakly, zhang2021weakly}.
For a given setting, all methods share the same labels.

\paragraphcond{Implementation and Training.}
We train on 4 NVIDIA Tesla V100 for 100 epochs, using AdamW with 0.01 learning rate and batch size 16.
We balance losses with $\lambda_\text{foc}=1.0$,  $\lambda_\text{lov}=1.0$ and $\lambda_\text{nce}=0.1$ and use standard data augmentation as in~\cite{cortinhal2020salsanext}.
Because early prototypes are too noisy, we apply a 5 epochs warm-up without contrastive learning.

\vspace{-0.5em}
\subsection{Performance}\vspace{-0.2em}
\label{sec:exp-perf}

\paragraphcond{Quantitative Results.}
We report performance on SemKITTI\footnote{For fair comparison with our SalsaNext backbone on the \textit{hidden} test set of SemKITTI we also use the kNN post-processing~\cite{rangenet++} to mitigate the risk of overlapping points in projection-output having same 3D label.}~\cite{semantickitti} \textit{hidden} test set in Tab.~\ref{tab:perf_semkitti}, and on SemPOSS~\cite{pan2020semanticposs} and nuScenes-Lidarseg~\cite{nuScenes} using their {validation sets} both in Tab.~\ref{tab:nuScenes-semantic-poss}.
On all datasets, we consistently outperform other weakly-supervised baselines being projection-based networks like SalsaNext~\cite{cortinhal2020salsanext} or point-based networks like SQN~\cite{sqn}.
Specifically, comparing Ours and SalsaNext - our backbone - the mIoU is +5.6/+3.6 mIoU on SemKITTI (Tab.~\ref{tab:perf_semkitti}) with only 0.1\%/0.01\% annotations and +4.1/+3.7 on SemPOSS (Tab.~\ref{tab:nuScenes-semantic-poss}a).
On nuScenes (Tab.~\ref{tab:nuScenes-semantic-poss}b), we get +2.2 mIoU with 0.1\% labels but\edit{ get}{} -1.6\% in the 0.01\% labels case.
We argue the drop relates to nuScenes (24x sparser than SemKITTI) having 10 of the 16 classes with less than 20 labels in the 0.01\% setting.
Subsequently, our clustering fails to associate labels with our 20 prototypes per class and the latter act as noise.
Importantly, in Tab.~\ref{tab:perf_semkitti} note that despite the projection-based nature of our backbone we significantly outperform SQN, a point-based method. Furthermore, we even beat \textit{some} recent fully supervised baselines on each dataset despite the fact that we use 1000x or even 10000x fewer labels.

\noindent{}This demonstrates our ability to learn robust semantic embeddings with significantly less labels. The ablation in Tab.~\ref{tab:ablation-exp}, later discussed, even advocates that 1\% of labels is sufficient to perform on par with full supervision.

\begin{table}[t]
	\scriptsize
	\setlength{\tabcolsep}{0.0025\linewidth}
	\newcommand{\classfreq}[1]{{\tiny(\semkitfreq{#1}\%)}}  %
	\centering
	\begin{tabular}{c|c|c|c|c c c c c c c c c c c c c c c c c c c}
		\toprule
		\rotatebox{90}{Anno.(\%)}
		                       & Method
		                       & \rotatebox{90}{Projection-based}
		                       & \rotatebox{90}{mIoU (\%)}
		                       & \rotatebox{90}{\textcolor{road}{$\blacksquare$} road~\classfreq{road}}
		                       & \rotatebox{90}{\textcolor{sidewalk}{$\blacksquare$} sidewalk~\classfreq{sidewalk}}
		                       & \rotatebox{90}{\textcolor{parking}{$\blacksquare$} parking~\classfreq{parking}}
		                       & \rotatebox{90}{\textcolor{other-ground}{$\blacksquare$} other-grnd~\classfreq{otherground}}
		                       & \rotatebox{90}{\textcolor{building}{$\blacksquare$} building~\classfreq{building}}
		                       & \rotatebox{90}{\textcolor{car}{$\blacksquare$} car~\classfreq{car}}
		                       & \rotatebox{90}{\textcolor{truck}{$\blacksquare$} truck~\classfreq{truck}}
		                       & \rotatebox{90}{\textcolor{bicycle}{$\blacksquare$} bicycle~\classfreq{bicycle}}
		                       & \rotatebox{90}{\textcolor{motorcycle}{$\blacksquare$} m.cycle~\classfreq{motorcycle}}
		                       & \rotatebox{90}{\textcolor{other-vehicle}{$\blacksquare$} other-veh.~\classfreq{othervehicle}}
		                       & \rotatebox{90}{\textcolor{vegetation}{$\blacksquare$} vegetation~\classfreq{vegetation}}
		                       & \rotatebox{90}{\textcolor{trunk}{$\blacksquare$} trunk~\classfreq{trunk}}
		                       & \rotatebox{90}{\textcolor{terrain}{$\blacksquare$} terrain~\classfreq{terrain}}
		                       & \rotatebox{90}{\textcolor{person}{$\blacksquare$} person~\classfreq{person}}
		                       & \rotatebox{90}{\textcolor{bicyclist}{$\blacksquare$} bicyclist~\classfreq{bicyclist}}
		                       & \rotatebox{90}{\textcolor{motorcyclist}{$\blacksquare$} m.cyclist~\classfreq{motorcyclist}}
		                       & \rotatebox{90}{\textcolor{fence}{$\blacksquare$} fence~\classfreq{fence}}
		                       & \rotatebox{90}{\textcolor{pole}{$\blacksquare$} pole~\classfreq{pole}}
		                       & \rotatebox{90}{\textcolor{traffic-sign}{$\blacksquare$} traf.-sign~\classfreq{trafficsign}}                                                                                                                                                                                                                                                                                                                                                                                                                        \\

		\midrule

		\multirow{8}{*}{100 }  & TangentConv~\cite{tangentconv}                                                                 & \multirow{4}{*}{\xmark}     & 40.9             & 83.9             & 63.9             & 33.4             & 15.4             & 83.4             & 90.8             & 15.2             & 2.7              & 16.5          & 12.1             & 79.5             & 49.3             & 58.1             & 23.0             & 28.4             & 8.1              & 49.0          & 35.8             & 28.5             \\

		                       & RandLA-Net~\cite{randlanet}                                                                    &                             & 55.9             & 90.5             & 74.0             & 61.8             & 24.5             & 89.7             & 94.2             & 43.9             & 47.4             & 32.2          & 39.1             & 83.8             & 63.6             & 68.6             & 48.4             & 47.4             & 9.4              & 60.4          & 51.0             & 50.7             \\

		                       & SPVNAS~\cite{spvnas}                                                                           &                             & 67.0             & 90.2             & 75.4             & 67.6             & 21.8             & 91.6             & 97.2             & 56.6             & 50.6             & 50.4          & 58.0             & 86.1             & 73.4             & 71.0             & 67.4             & 71.0             & 50.3             & 66.9          & 64.3             & 67.3             \\

		                       & $\mathrm{(AF)^2S3Net}$~\cite{2s3net}                                                           &                             & 69.7             & 91.3             & 72.5             & 68.8             & 53.5             & 87.9             & 94.5             & 39.2             & 65.4             & 86.8          & 41.1             & 70.2             & 68.5             & 53.7             & 80.7             & 74.3             & 74.3             & 63.2          & 61.5             & 71.0             \\

		\cmidrule{2-23}
		                       & DarkNet53Seg~\cite{semantickitti}                                                              & \multirow{4}{*}{\checkmark} & 49.9             & 91.8             & 74.6             & 64.8             & 27.9             & 84.1             & 86.4             & 25.5             & 24.5             & 32.7          & 22.6             & 78.3             & 50.1             & 64.0             & 36.2             & 33.6             & 4.7              & 55.0          & 38.9             & 52.2             \\

		                       & RangeNet53++~\cite{rangenet++}                                                                 &                             & 52.2             & 91.8             & 75.2             & 65.0             & 27.8             & 87.4             & 91.4             & 25.7             & 25.7             & 34.4          & 23.0             & 80.5             & 55.1             & 64.6             & 38.3             & 38.8             & 4.8              & 58.6          & 47.9             & 55.9             \\

		                       & SqueezeSegV3~\cite{squeezesegv3}                                                               &                             & 55.9             & 91.7             & 74.8             & 63.4             & 26.4             & 89.0             & 92.5             & 29.6             & 38.7             & 36.5          & 33.0             & 82.0             & 58.7             & 65.4             & 45.6             & 46.2             & 20.1             & 59.4          & 49.6             & 58.9             \\

		                       & SalsaNext~\cite{cortinhal2020salsanext}                                                        &                             & 59.5             & 91.7             & 75.8             & 63.7             & 29.1             & 90.2             & 91.9             & 38.9             & 48.3             & 38.6          & 31.9             & 81.8             & 63.6             & 66.5             & 60.2             & 59.0             & 19.4             & 64.2          & 54.3             & 62.1             \\

		\midrule
		\multirow{3}{*}{0.1 }  & SQN~\cite{sqn}                                                                                 & \xmark                      & \underline{50.8} & \textbf{90.5}    & \textbf{72.9}    & \underline{56.8} & \underline{19.1} & 84.8             & \textbf{92.1}    & \underline{36.7} & \textbf{39.3}    & 30.1          & 26.0             & \textbf{80.8}    & \underline{59.1} & \textbf{67.0}    & 36.4             & 25.3             & 7.2              & 53.3          & \underline{44.5} & \underline{44.0} \\
		                       & SalsaNext~\cite{cortinhal2020salsanext}                                                        & \multirow{2}{*}{\checkmark} & 50.1
		                       & 87.2                                                                                           & 66.0                        & 51.6             & 18.4             & \underline{86.2} & 88.2             & 31.6             & 25.3             & \underline{29.9} & \underline{31.8} & 77.0             & 59.0          & 60.1             & \underline{44.6} & \underline{44.2} & \underline{15.6} & \underline{54.9} & 39.6             & 41.4                                                                   \\
		                       & Ours                                                                                           &                             & \textbf{55.7}    & \underline{88.4} & \underline{68.2} & \textbf{57.6}    & \textbf{23.4}    & \textbf{87.7}    & \underline{89.7} & \textbf{41.0}    & \underline{36.2} & \textbf{36.3} & \textbf{38.3}    & \underline{79.1} & \textbf{62.3}    & \underline{60.2} & \textbf{52.6}    & \textbf{45.0}    & \textbf{24.6}    & \textbf{58.1} & \textbf{49.4}    & \textbf{59.9}    \\

		\midrule
		\multirow{3}{*}{0.01 } & SQN~\cite{sqn}                                                                                 & \xmark                      & 39.1             & \underline{86.6} & \underline{66.4} & 43.0             & \underline{16.9} & 80.0             & \underline{85.5} & 12.9             & 4.0              & 1.4           & \underline{18.4} & 72.7             & \underline{49.6} & 58.8             & 16.9             & 22.3             & \underline{4.3}  & 42.3          & 31.7             & 16.6             \\
		                       & SalsaNext~\cite{cortinhal2020salsanext}                                                        & \multirow{2}{*}{\checkmark} & \underline{42.6}                                                                                                                                                                                                                                                                                                                                                                    %
		                       & 86.0                                                                                           & 65.4                        & \underline{43.7} & 14.5             & \underline{85.1} & 83.1             & \underline{19.5} & \underline{21.9} & \underline{14.9} & 17.6             & \underline{73.8} & 45.9          & \underline{59.7} & \underline{29.4} & \underline{23.6} & \underline{4.2}  & \underline{50.7} & \underline{32.6} & \underline{38.1}                                                       \\

		                       & Ours                                                                                           &                             & \textbf{46.2}    & \textbf{88.4}    & \textbf{68.0}    & \textbf{52.7}    & \textbf{17.4}    & \textbf{87.0}    & \textbf{87.3}    & \textbf{28.5}    & \textbf{25.3}    & \textbf{16.0} & \textbf{23.6}    & \textbf{79.8}    & \textbf{55.5}    & \textbf{62.8}    & \underline{28.6} & \textbf{25.3}    & 1.5              & \textbf{55.0} & \textbf{40.3}    & \textbf{43.9}    \\
		\bottomrule
	\end{tabular}
	\vskip 0.02 in
	\caption{On SemKITTI~\cite{semantickitti} \textit{hidden} test set, our method outperforms all weakly-supervised baselines with either 0.1\% or 0.01\% labels and \textit{some} fully supervised methods.}
	\label{tab:perf_semkitti}
	\vskip 1.0em

	\centering
	\tablesize
	\setlength{\tabcolsep}{0.004\textwidth}
	\begin{tabular}{P{0.5\linewidth} P{0.5\linewidth}}
		\begin{tabular}{lcc}
			\toprule
			Method                                  & Anno.(\%)             & mIoU\%        \\
			\midrule

			RandLA-Net~\cite{randlanet}             & \multirow{3}{*}{100}  & 53.5          \\

			KPConv~\cite{thomas2019kpconv}          &                       & 55.2          \\

			JS3C-Net~\cite{js3c-net}                &                       & 60.2          \\

			\midrule
			SqueezeSegV2~\cite{squeezesegv2}        & \multirow{2}{*}{100}  & 29.8          \\

			SalsaNext~\cite{cortinhal2020salsanext} &                       & 45.0          \\
			\midrule
			SalsaNext~\cite{cortinhal2020salsanext} & \multirow{2}{*}{0.1}  & 38.9          \\
			Ours                                    &                       & \textbf{43.0} \\
			\midrule
			SalsaNext~\cite{cortinhal2020salsanext} & \multirow{2}{*}{0.01} & 27.4          \\
			Ours                                    &                       & \textbf{31.1} \\
			\bottomrule
		\end{tabular} &
		\begin{tabular}{lcc}
			\toprule
			Method                                  & Anno.(\%)             & mIoU(\%)      \\
			\midrule

			RandLA-Net~\cite{randlanet}             & \multirow{3}{*}{100}  & 53.5          \\

			PolarNet~\cite{polarnet}                &                       & 72.2          \\
			Cylinder3D~\cite{cylinder3d}            &                       & 76.1          \\

			\midrule
			RangeNet++~\cite{rangenet++}            & \multirow{2}{*}{100}  & 65.5          \\

			SalsaNext~\cite{cortinhal2020salsanext} &                       & 72.2          \\
			\midrule

			SalsaNext~\cite{cortinhal2020salsanext} & \multirow{2}{*}{0.1}  & 56.5          \\

			{Ours}                                  &                       & \textbf{58.7} \\
			\midrule

			SalsaNext~\cite{cortinhal2020salsanext} & \multirow{2}{*}{0.01} & \textbf{44.5}
			\\

			{Ours}                                  &                       & 42.9          %
			\\
			\bottomrule
		\end{tabular} \\
		(a) SemPOSS                                                                        & (b) nuScenes
	\end{tabular}
	\caption{Performance on SemPOSS~\cite{pan2020semanticposs} (a) and nuScenes~\cite{nuScenes} (b) validation sets.
	}
	\label{tab:nuScenes-semantic-poss}\vspace{-0.2em}

\end{table}

\begin{figure*}
	\centering
	\includegraphics[width=0.95\linewidth]{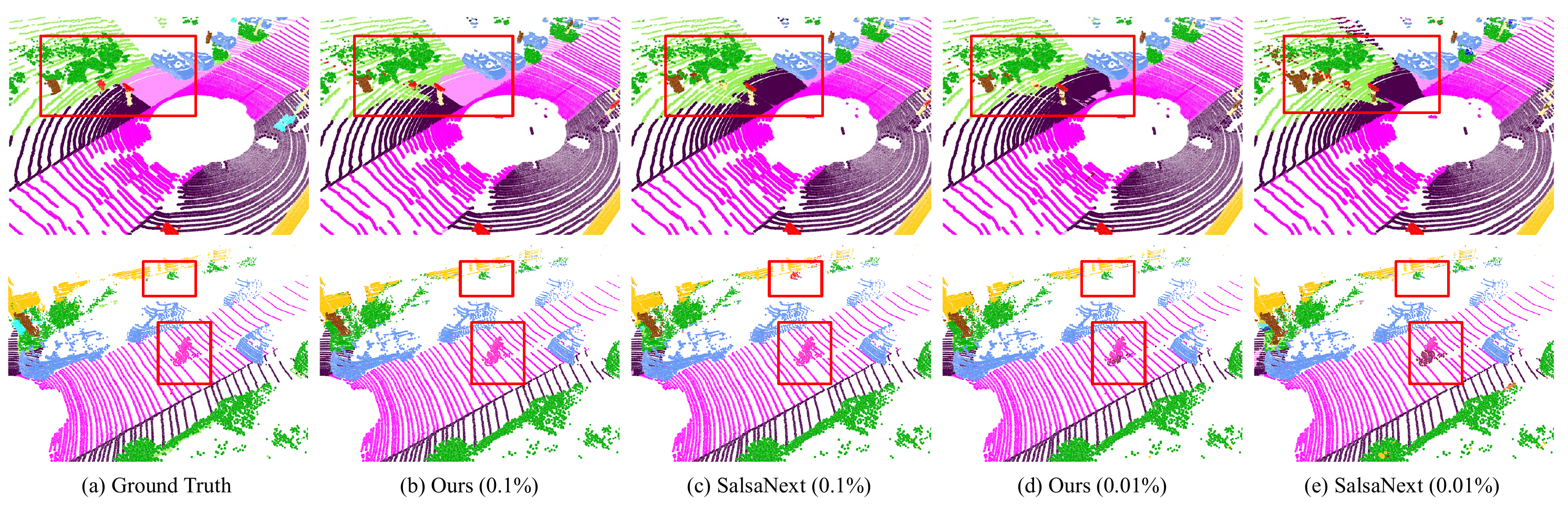}
	\\[-0.2em]
	\scriptsize
	\textcolor{bicycle}{$\blacksquare$}bicycle~
	\textcolor{car}{$\blacksquare$}car~
	\textcolor{motorcycle}{$\blacksquare$}motorcycle~
	\textcolor{truck}{$\blacksquare$}truck~
	\textcolor{other-vehicle}{$\blacksquare$}other vehicle~
	\textcolor{person}{$\blacksquare$}person~
	\textcolor{bicyclist}{$\blacksquare$}bicyclist~
	\textcolor{motorcyclist}{$\blacksquare$}motorcyclist~
	\textcolor{road}{$\blacksquare$}road~
	\textcolor{parking}{$\blacksquare$}parking~
	\\
	\scriptsize
	\textcolor{sidewalk}{$\blacksquare$}sidewalk~
	\textcolor{other-ground}{$\blacksquare$}other ground~
	\textcolor{building}{$\blacksquare$}building~
	\textcolor{fence}{$\blacksquare$}fence~
	\textcolor{vegetation}{$\blacksquare$}vegetation~
	\textcolor{trunk}{$\blacksquare$}trunk~
	\textcolor{terrain}{$\blacksquare$}terrain~
	\textcolor{pole}{$\blacksquare$}pole~
	\textcolor{traffic-sign}{$\blacksquare$}traffic sign
	\\[-1em]
	\caption{Sampled predictions on SemKITTI~\cite{semantickitti} validation set.}
	\label{fig:semantic-kitti-visualization}
	\vspace{-0.5em}
\end{figure*}

\paragraphcond{Qualitative Results.}
Fig.~\ref{fig:semantic-kitti-visualization} shows qualitative outputs on SemKITTI~\cite{semantickitti} validation set, along with ground truth and SalsaNext~\cite{cortinhal2020salsanext} baseline given the absence of public implementation for SQN~\cite{sqn}.
In both 0.1\% and 0.01\% settings, our method surpasses SalsaNext, especially on thin objects. 
More qualitative results are in the Appendix~\ref{sec:addresults}. %

\vspace{-0.5em}
\subsection{Ablation Study}
\vspace{-0.2em}
\label{sec:exp-ablation-study}
We first vary the backbone used and then ablate our method. Unless otherwise mentioned, we employ SalsaNet backbone with 0.1\% annotation evaluated on SemKITTI val. set.

\paragraphcond{Choice of Backbones.}
To demonstrate our approach is architecture-agnostic, Tab.~\ref{tab:backbones} shows experiments on the val. sets of SemKITTI~\cite{semantickitti} and SemPOSS~\cite{pan2020semanticposs} with various backbones, namely: RangeNet-21~\cite{rangenet++}, SqueezeSegV3-21~\cite{squeezesegv3}, and SalsaNext~\cite{cortinhal2020salsanext} 
(details in Appendix~\ref{sec:implementation-backbone}).
For all backbones and datasets, we boost results significantly w.r.t. to the original segmentation backbone, with mIoU difference of $[+3.8,+6.3]$. The results also advocate for our choice of SalsaNext backbone which performs best over all three.

\paragraphcond{Architectural Components.}
Tab.~\ref{tab:ablation-exp}a shows that all of our designs contribute to the best results.
More in depth, `w/o contrast module' shows that the contrastive module (Sec.~\ref{sec:pix2proto_contrastive_learning}) contributes to the biggest gain.
To evaluate the benefit of our anchor sampling (Sec.~\ref{sec:achors_sampling}), `w/o anchor sampling' instead uses all pseudo-labels as anchors showing our strategy is indeed beneficial.
In `w/o prototype' we replace our 20 compact prototypes (Sec.~\ref{sec:proto_mem_bank}) with~5,000 pixels as in~\cite{cross_img_contrast}, showing that our method is lighter and more efficient.
Finally, Focal~loss~(Eq.~\ref{eq:focloss}) and Lovasz loss (Eq.~\ref{eq:lovloss}) are better combined as the two pursue slightly different objectives since Focal loss mitigates imbalance while Lovasz maximizes IoU.

\paragraphcond{Memory Bank.} We now compare other types of memory banks of various per-class sizes.
From Tab.~\ref{tab:ablation-exp}b, we compare against two variations from~\cite{cross_img_contrast}:
`Pixel' stores abundant random pixel-wise embeddings and `Region' which instead stores frame-wise pooled features of semantically consistent regions.
Compared to both, our method is more efficient and lighter.
The combination of `Pixel+Region' using 10,000 bank size, proves to be better but still gets outperformed with only 20 of our prototypes. Finally, we also compare against prototypes from mini-batch embeddings~\cite{local-aggregation} -- which logically perform worse since they fail at capturing the complete dataset knowledge.
In a nutshell, Tab.~\ref{tab:ablation-exp} reveals that more stable contrastive targets -- like regions or prototypes -- are better and that our momentum update mechanism preserves the dataset context thus reaching better performance.

\begin{table}
	\centering
	\tablesize
	\setlength{\tabcolsep}{2pt}
	\begin{tabular}{crlrl}
		\toprule
		\multirow{2}{*}{Method}                 & \multicolumn{2}{c}{SemPOSS~\cite{pan2020semanticposs}} & \multicolumn{2}{c}{SemKITTI~\cite{semantickitti}}                                                                   \\
		                                        & \multicolumn{2}{c}{mIoU~(\%)}                               & \multicolumn{2}{c}{mIoU~(\%)}                                                                                            \\
		\midrule

		Rangenet-21~\cite{rangenet++}           & 25.1                                                        & \reflectbox{\rotatebox[origin=c]{270}{$\drsh$}}        & 40.7          & \reflectbox{\rotatebox[origin=c]{270}{$\drsh$}} \\
		{Ours (Rangenet-21)}                    & \textbf{28.9}                                               & {\scriptsize +3.8}                                     & \textbf{44.5} & {\scriptsize +3.8}                              \\
		\cmidrule{1-5}
		SqueezeSegV3-21~\cite{squeezesegv3}     & 30.4                                                        & \reflectbox{\rotatebox[origin=c]{270}{$\drsh$}}        & 42.5          & \reflectbox{\rotatebox[origin=c]{270}{$\drsh$}} \\
		{Ours (SqueezeSegV3-21)}                & \textbf{36.7}                                               & {\scriptsize +6.3}                                     & \textbf{48.5} & {\scriptsize +6.0}                              \\
		\cmidrule{1-5}
		SalsaNext~\cite{cortinhal2020salsanext} & {38.9}                                                      & \reflectbox{\rotatebox[origin=c]{270}{$\drsh$}}        & {52.4}        & \reflectbox{\rotatebox[origin=c]{270}{$\drsh$}} \\
		{Ours (SalsaNext)}                      & \textbf{43.0}                                               & {\scriptsize +4.1}                                     & \textbf{57.6} & {\scriptsize +5.2}                              \\

		\bottomrule
	\end{tabular}
	\vskip 0.05 in
	\caption{COARSE3D with different backbones using 0.01\% annotation (val. set).}
	\label{tab:backbones}
	\vspace{-0.5em}
\end{table}

\begin{table}
	\centering
	\tablesize
	\setlength{\tabcolsep}{0.004\textwidth}
	\begin{tabular}{P{0.35\linewidth} P{0.4\linewidth} P{0.23\linewidth}}
		\begin{tabular}{cc}

			\toprule
			Method                 & mIoU (\%)         \\
			\midrule
			Ours                   & \textbf{57.57}    \\
			w/o contrast module    & 55.44             \\
			w/o anchors sampling   & \underline{56.32} \\
			w/o prototype (5k pxl) & 56.10             \\ %
			w/o voxel propagation  & 56.26             \\
			\midrule
			w/o Focal loss         & 42.41             \\
			w/o Lov{\'a}sz loss        & 56.10             \\
			\bottomrule
		\end{tabular} &
		\begin{tabular}{ccc}
			\toprule
			Memory bank            & Bank size & mIoU (\%)         \\
			\midrule
			$w/o$ contrast         & -         & 55.44             \\
			\midrule
			Pixel                  & 5,000     & 56.10             \\
			Region                 & 5,000     & 56.66             \\
			Pixel + Region         & 10,000    & \underline{56.79} \\
			\midrule
			Prototype (mini-batch) & 20        & 56.46             \\
			Prototype (ours)       & 20        & \textbf{57.57}    \\
			\bottomrule
		\end{tabular}                       &
		\begin{tabular}{cc}

			\toprule

			\# Prototype~ & mIoU (\%)         \\
			\midrule
			1             & 55.47             \\
			10            & 56.15             \\
			20            & \textbf{57.57}    \\
			50            & 56.18             \\
			100           & \underline{56.89} \\
			\bottomrule
		\end{tabular}                                                                                         \\
		(a) Method ablation                                                             & (b) Memory bank & (c) Prototypes number
	\end{tabular}

	\vskip 0.1 in

	\centering
	\tablesize
	\setlength{\tabcolsep}{0.004\textwidth}
	\begin{tabular}{P{0.3\linewidth} P{0.3\linewidth} P{0.3\linewidth}}
		\begin{tabular}{ccc}
			\toprule

			Strategy           & Basis                    & mIoU(\%)          \\
			\midrule
			All                & \multirow{2}{*}{No}      & 56.32             \\
			Random K           &                          & 56.06             \\
			\midrule
			Hard-threshold     & \multirow{3}{*}{Softmax} & \underline{57.09}
			\\
			Top-K              &                          & 56.11
			\\
			Probability        &                          & 56.93             \\
			\midrule
			Hard-threshold     & \multirow{3}{*}{Entropy} & 56.38             \\
			Top-K              &                          & 56.30             \\
			Probability (Ours) &                          & \textbf{57.57}    \\
			\bottomrule
		\end{tabular}
		                                          &
		\begin{tabular}{cc}
			\toprule
			Strategy   & mIoU(\%)            \\
			\midrule
			Easy       & 56.15               \\
			Hard       & $\underline{57.00}$ \\
			Semi Hard  & 56.59               \\
			Random     & 56.41               \\
			All (Ours) & \textbf{57.57}      \\
			\bottomrule
		\end{tabular} &

		\begin{tabular}{c|cc}
			\toprule
			\multirow{2}{*}{Anno.} & \multicolumn{2}{c}{mIoU (\%)}                   \\
			                       & SalsaNext~\cite{cortinhal2020salsanext} & Ours  \\
			\midrule
			0.001\%                & 30.39                                   & 31.69 \\
			0.01\%                 & 44.00                                   & 47.13 \\
			0.1\%                  & 52.43                                   & 56.61 \\
			1\%                    & 56.16                                   & 58.30 \\
			100\%                  & 56.44                                   & 58.39 \\
			\bottomrule
		\end{tabular}

		\\
		(d) Anchor sampling                       & (e) Key sampling & (f) Annotation \\
	\end{tabular}
	\vskip 0.02 in
	\caption{Ablation study results on SemKITTI\cite{semantickitti} validation set.}
	\label{tab:ablation-exp}
	\vspace{-0.7em}
\end{table}

\paragraphcond{Number of Prototypes.}
Tab.~\ref{tab:ablation-exp}c studies the impact of varying the number of per class prototypes from 1 to 100.
With only 1 prototype, each class uses a single high-dimensional embedding which is also referred to as class-wise memory bank~\cite{alonso2021semi}.
Altogether, 20 prototypes are optimal among tested choices since fewer or more prototypes degrade performance.

\paragraphcond{Choice of Anchors.}
We study the choice of anchors in Tab.~\ref{tab:ablation-exp}d, by varying the basic function \edit{that is }{}used for sampling (col `Basis') and the strategy for sampling segmentation features (col `Strategy').
Without any basis function, we tried using all pseudo labels predictions `All' or a `Random X' number of X pixels. Results show that more support from pseudo labels boosts performance but the strategy is suboptimal given the prediction inaccuracies.
To mitigate this, we evaluate the predictions quality with a basis function: either a `Softmax' as~\cite{defense-pseudo-label} or our `Entropy' proposal (Sec.~\ref{sec:achors_sampling}).
For both `Softmax' and `Entropy' basis, we consider sampling with a finetuned `hard-threshold', the `Top-X' elements, or selection from a random `Probability' sampling.
Our `Entropy' driven probability sampling performs best.
We relate these results to the fact that Shannon entropy stores joint information distribution for all class predictions, which makes it more insightful for a proper anchor selection.

\paragraphcond{Choice of Keys.}
Prior research~\cite{cross_img_contrast,hard-negative-sample} show the benefit of using smart key sampling strategies which differs to our use of all keys. In Tab.~\ref{tab:ablation-exp}e we compare the common key sampling strategies being\edit{ either}{} `easy', `hard', `semi-hard', `random' or `All'.
Except for the `All' strategy, we consider 100 prototypes and use the key sampling mentioned to select 20 of these prototypes as keys.
In short, `easy'/`hard' indicates the complexity for anchors to learn with such keys. We refer to~\cite{cross_img_contrast} for details.
Intuitively, using semi-hard or hard keys encourage the network to learn more robust representations. However, we denote that using all keys performs in fact better with our prototypes. We conjecture that this relates to the prototype capturing a more robust dataset context which is beneficial to preserve.

\paragraphcond{Effect of Annotation.} In Tab.~\ref{tab:ablation-exp}f we vary the annotation ratio from 100\% (full supervision) to only 0.001\% (roughly 1 point label per frame), showing our robustness to the most extreme cases -- despite a drastic mIoU drop for 0.001\% annotation due to rare classes being absent from labels. In fact, reporting for the latter only mIoU for `classes in labels', SalsaNext and Ours get respectively, 45.50\% and 47.61\% which again demonstrates robustness.
We further study the effect of random label in Appendix~\ref{sec:annotations}, 
showing in a nutshell that variances of 0.1\%/0.01\% settings are $\pm$0.93/$\pm$0.32 -- far smaller than the baseline gaps (+4.18/+3.13).

\vspace{-0.5em}
\section{Conclusion}\vspace{-0.3em}
In this paper, we present a weakly supervised LiDAR point cloud semantic segmentation framework. Specifically, we develop a compact class-prototype contrastive learning scheme based on online clustered embeddings and use this prototype as the key to pulling entropy-driven sampled anchors. 
Extensive experiments on three projection-based backbones and three real-world datasets demonstrate the effectiveness of our method.

\begin{spacing}{0.5}
\footnotesize
\paragraphcond{Acknowledgement.} 
 Rong Li was supported by the SMIL lab of South China University of Technolog, received support and advices from Prof. Mingkui Tan, Prof. Caixia Li and Zhuangwei Zhuang. Rong Li was also partly supported by Key-Area Research and Development Program Guangdong Province (2019B010155001). Inria members were partly funded by French project SIGHT (ANR-20-CE23-0016). This work was performed using HPC resources from GENCI–IDRIS (Grant 2021-AD011012808 and 2022-AD011012808R1).
\end{spacing}


\appendix

We study the effect of annotation sampling in Appendix~\ref{sec:annotations} and report implementation details with new ablation in Appendix~\ref{sec:implementation} and more qualitative results in Appendix~\ref{sec:addresults}.

\vspace{-0.5em}
\section{Effect of Annotations}\vspace{-0.5em}
\label{sec:annotations}
In weak supervision settings, changing the set of labelled points impact performance. We now study more in depth the effect of sampled labels by randomly resampling labels 3 times on 2 datasets. As it appears in Tab.~\ref{tab:labresampling}, our method is relatively stable (\textit{i.e.}, $\text{std}{<}1$). We also measure the gap between random annotations and those of human by asking 2~operators to annotate the entire SemPOSS, labeling roughly 0.01\% points per frame.
Again, from Tab.~\ref{tab:labresampling} our `human' labels are within 3 std of the mean 0.01\% performance (\textit{i.e.}, 29.27 vs 31.48{\footnotesize$\pm0.43$}).

\begin{table}[h]
	\scriptsize
	\centering
	\setlength{\tabcolsep}{0.014\linewidth}
	\renewcommand{\arraystretch}{0.8}

	\newcommand{\meanstd}[2]{{\scriptsize#1{\scriptsize{}$\pm$#2}}}
	\begin{tabular}{c|cc||cc|c}
		\toprule
		      & \multicolumn{2}{c||}{\textbf{SemKITTI~\cite{semantickitti}}} & \multicolumn{3}{c}{\textbf{SemPOSS~\cite{pan2020semanticposs}}}                                                                   \\
		Anno. & 0.10\%                                                            & 0.01\%                                                               & 0.10\%                & 0.01\%                & \makecell{human \\[-0.3em]{\scriptsize ($\approx{}$0.01\%)}}\\ \midrule
		run 1 & 57.57                                                             & 47.35                                                                & 43.00                 & 31.10                 & 29.27           \\
		run 2 & 56.54                                                             & 47.28                                                                & 42.88                 & 31.95                 & -               \\
		run 3 & 55.71                                                             & 46.76                                                                & 42.47                 & 31.38                 & -               \\ \midrule
		all   & \meanstd{56.61}{0.93}                                             & \meanstd{47.13}{0.32}                                                & \meanstd{42.78}{0.28} & \meanstd{31.48}{0.43} & -               \\
		\bottomrule
	\end{tabular}\vspace{1em}
	\caption{Effect of annotation sampling on SemKITTI and SemPOSS.}
	\label{tab:labresampling}
\end{table}

\vspace{-0.5em}
\section{Implementation Details}\vspace{-0.2em}
\label{sec:implementation}
\subsection{Label Voxel Propagation}
\label{sec:implementation-voxel}
We replicate SQN~\cite{sqn} and apply their random grid downsampling with 0.06 voxel size.
Our trivial scheme simply propagates existing labels to all points within the same voxel -- thus densifying the labels at no extra labelling cost.
In the extremely rare case of conflicting labels within a single voxel (e.g. a voxel having two labelled points with different classes), the voxel label will be randomly assigned.

We evaluate the effect of this voxel propagation scheme using the SalsaNext backbone on SemKITTI val. set in the 0.1\% annotation setting.
With/without our scheme we get \textbf{57.57}/56.26 mIoU. Since baselines do not use our label propagation, it is important to note that the mIoU gap obtained (+1.31 mIoU) is smaller than the gap with the original SalsaNext (+5.14, cf. main paper Tab.~3f) in the same 0.1\% setting. This advocates that our method only \textit{partly} benefits from our voxel propagation scheme and performs best thanks to our overall contrastive learning strategy.

\subsection{Segmentation Backbones}
\label{sec:implementation-backbone}

\paragraphcond{SalsaNext~\cite{cortinhal2020salsanext}.}
We use the official implementation\footnote{\url{https://github.com/TiagoCortinhal/SalsaNext}} for the SemKITTI dataset\cite{semantickitti} and applied our best effort to fairly re-implement it for nuScenes~\cite{nuScenes} and SemPOSS~\cite{pan2020semanticposs}.
Specific to SemPOSS\cite{pan2020semanticposs}, we used an input padding to get a compatible size.
When trained with our method, we finetune our contrastive learning hyperparameters.

\paragraphcond{\textbf{SqueezeSegV3~\cite{squeezesegv3}.}}
We use the lighter \emph{SqueezeSegV3-21} from the official implementation\footnote{\url{https://github.com/chenfengxu714/SqueezeSegV3}}.
To boost performance on weakly supervised tasks, we replaced the multi-layer cross-entropy loss -- improper for sparse weak labels due to its downsampling --, with normal cross-entropy loss.
When trained with our method, we simply use the contrastive learning hyperparameter found with SalsaNext.

\paragraphcond{RangeNet++~\cite{rangenet++}.}
We use the ligher \emph{RangeNet-21} from the official implementation\footnote{\url{https://github.com/PRBonn/lidar-bonnetal}}.
Inputs are pad to get compatible size.
When trained with our method, we simply use the contrastive learning hyperparameter found with SalsaNext.

\vspace{-0.5em}
\section{Additional Results}\vspace{-0.2em}
\label{sec:addresults}

\begin{figure*}
	\centering
	\includegraphics[width=0.925\linewidth]{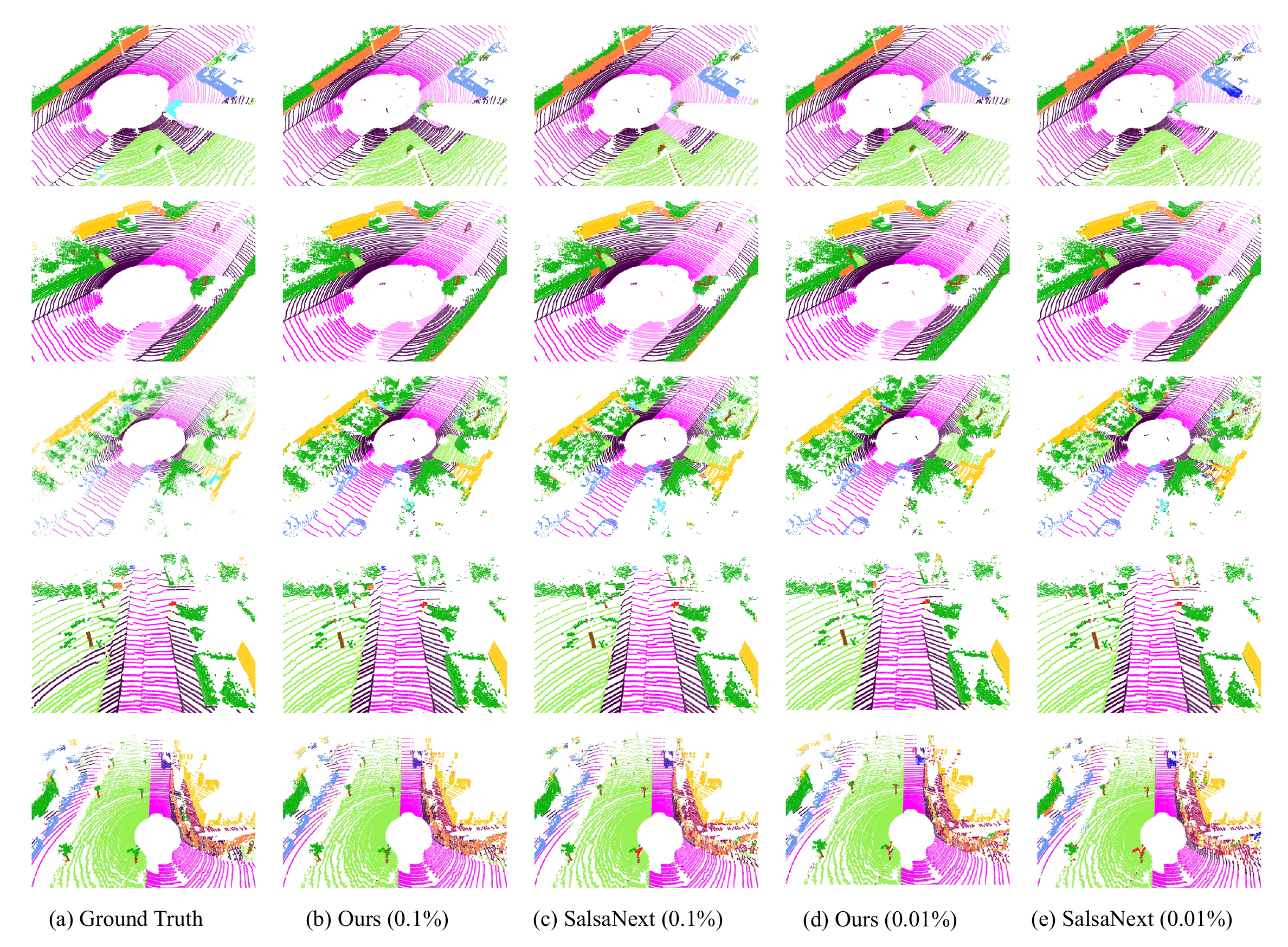}
	\\[-0.1em]

	\scriptsize
	\textcolor{bicycle}{$\blacksquare$}bicycle~
	\textcolor{car}{$\blacksquare$}car~
	\textcolor{motorcycle}{$\blacksquare$}motorcycle~
	\textcolor{truck}{$\blacksquare$}truck~
	\textcolor{other-vehicle}{$\blacksquare$}other vehicle~
	\textcolor{person}{$\blacksquare$}person~
	\textcolor{bicyclist}{$\blacksquare$}bicyclist~
	\textcolor{motorcyclist}{$\blacksquare$}motorcyclist~
	\textcolor{road}{$\blacksquare$}road~
	\textcolor{parking}{$\blacksquare$}parking~
	\\
	\scriptsize
	\textcolor{sidewalk}{$\blacksquare$}sidewalk~
	\textcolor{other-ground}{$\blacksquare$}other ground~
	\textcolor{building}{$\blacksquare$}building~
	\textcolor{fence}{$\blacksquare$}fence~
	\textcolor{vegetation}{$\blacksquare$}vegetation~
	\textcolor{trunk}{$\blacksquare$}trunk~
	\textcolor{terrain}{$\blacksquare$}terrain~
	\textcolor{pole}{$\blacksquare$}pole~
	\textcolor{traffic-sign}{$\blacksquare$}traffic sign
	\\[1em]

	\caption{Additional qualitative results on SemKITTI~\cite{semantickitti}}\vspace{-1.5em}
	\label{fig:semantic-semkitti-visualization}

\end{figure*}

\begin{figure*}
	\centering
	\includegraphics[width=0.925\linewidth]{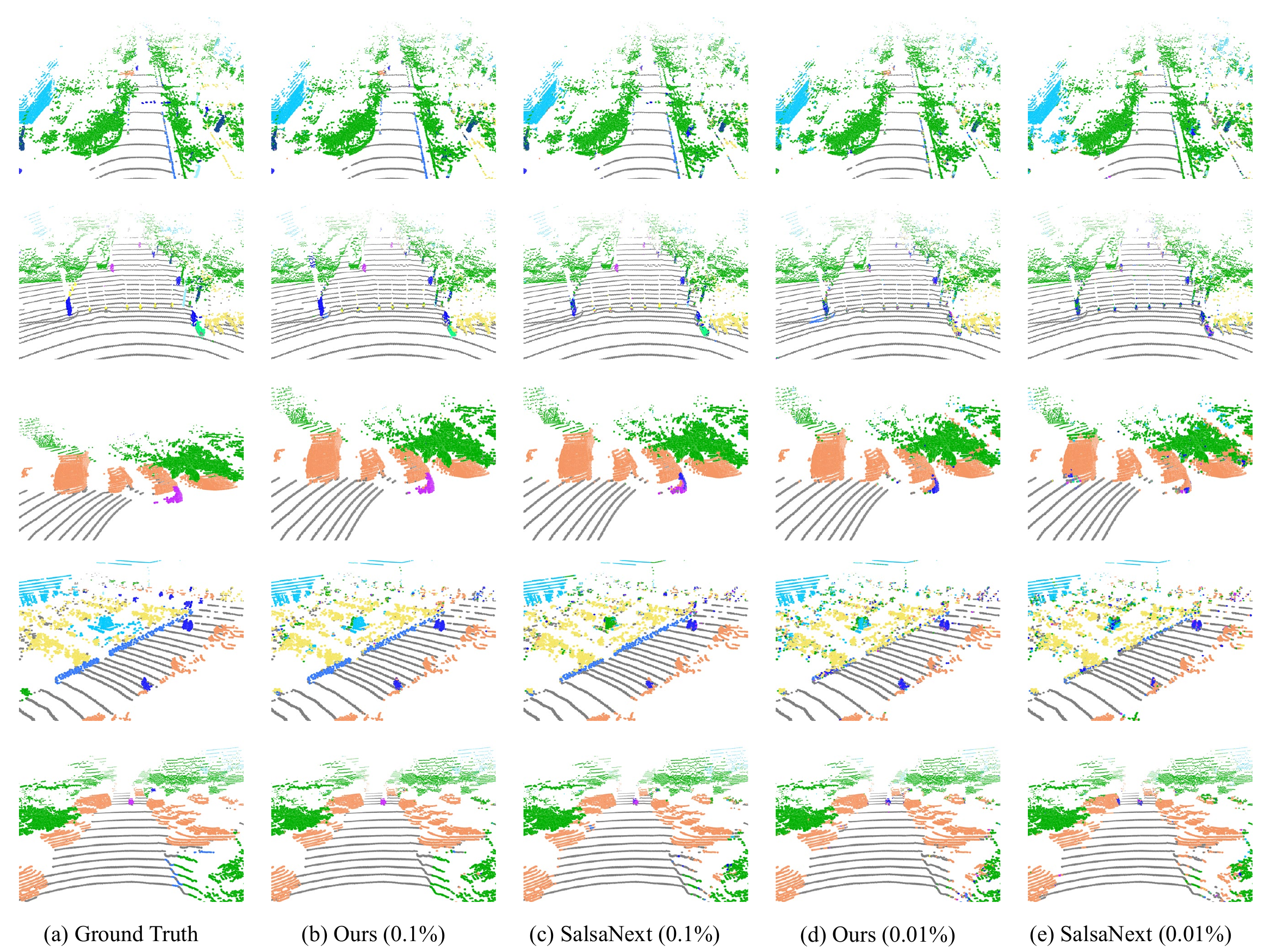}
	\\[-0.1em]

	\scriptsize
	\textcolor{poss01}{$\blacksquare$}people~
	\textcolor{poss02}{$\blacksquare$}rider~
	\textcolor{poss03}{$\blacksquare$}car~
	\textcolor{poss04}{$\blacksquare$}trunk~
	\textcolor{poss05}{$\blacksquare$}plants~
	\textcolor{poss06}{$\blacksquare$}traffic-sign~
	\\
	\scriptsize
	\textcolor{poss07}{$\blacksquare$}pole~
	\textcolor{poss08}{$\blacksquare$}trashcan~
	\textcolor{poss09}{$\blacksquare$}building~
	\textcolor{poss10}{$\blacksquare$}cone/stone~
	\textcolor{poss11}{$\blacksquare$}fence~
	\textcolor{poss12}{$\blacksquare$}bike~
	\textcolor{poss13}{$\blacksquare$}road~
	\caption{Qualitative results on SemPOSS~\cite{pan2020semanticposs}}\vspace{-1.5em}
	\label{fig:semantic-semposs-visualization}

\end{figure*}

\begin{figure*}
	\centering
	\includegraphics[width=0.925\linewidth]{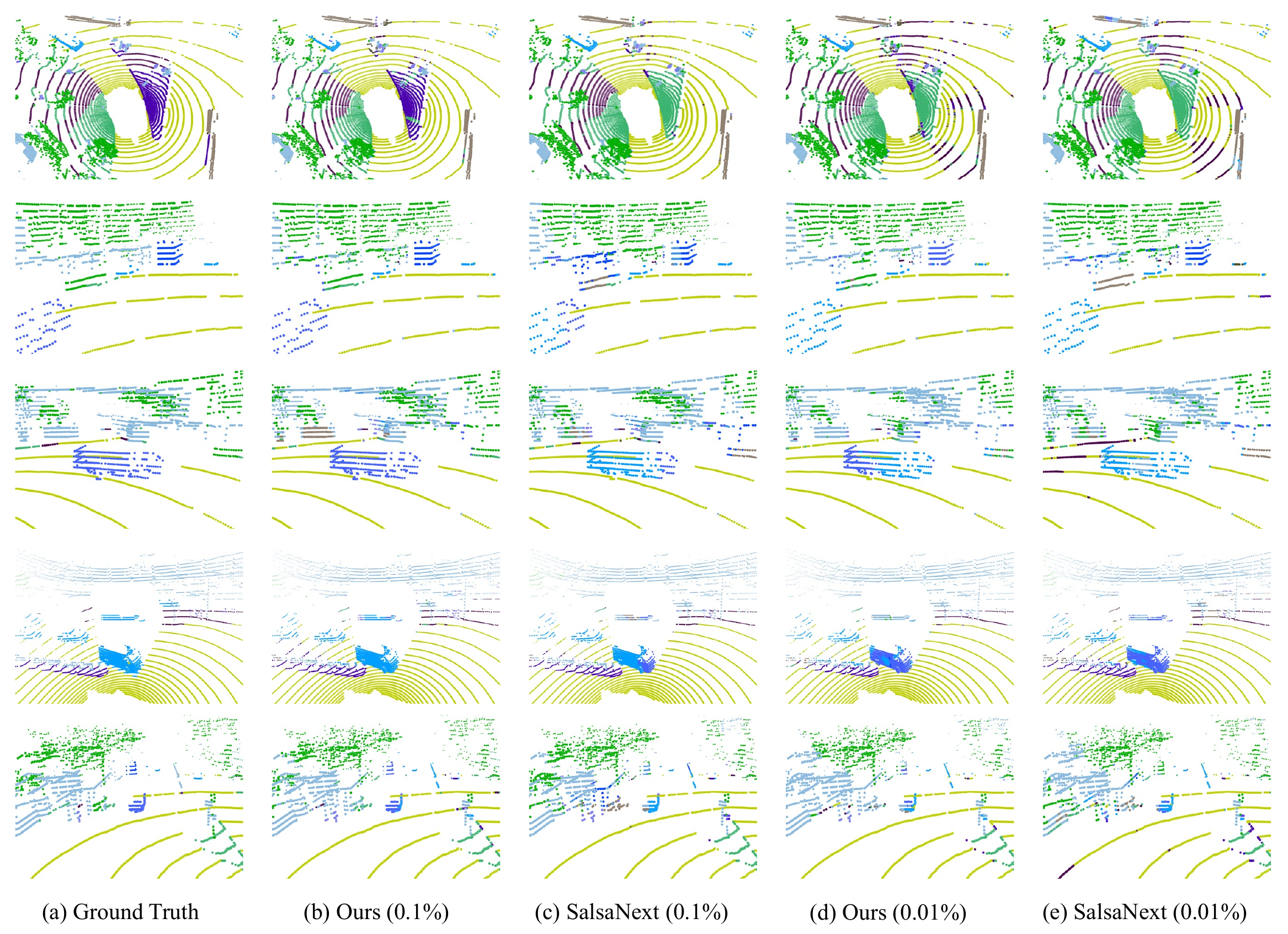}
	\\[-0.1em]

	\scriptsize
	\textcolor{nus01}{$\blacksquare$}barrier~
	\textcolor{nus02}{$\blacksquare$}bicycle~
	\textcolor{nus03}{$\blacksquare$}bus~
	\textcolor{nus04}{$\blacksquare$}car~
	\textcolor{nus05}{$\blacksquare$}construction vehicle~
	\textcolor{nus06}{$\blacksquare$}motorcycle~
	\textcolor{nus07}{$\blacksquare$}pedestrian~
	\textcolor{nus08}{$\blacksquare$}traffic cone~
	\textcolor{nus09}{$\blacksquare$}trailer~
	\\
	\scriptsize
	\textcolor{nus10}{$\blacksquare$}truck~
	\textcolor{nus11}{$\blacksquare$}driveable surface~
	\textcolor{nus12}{$\blacksquare$}other flat~
	\textcolor{nus13}{$\blacksquare$}sidewalk~
	\textcolor{nus14}{$\blacksquare$}terrain~
	\textcolor{nus15}{$\blacksquare$}manmade~
	\textcolor{nus16}{$\blacksquare$}vegetation~
	\caption{Qualitative results on nuScenes~\cite{nuScenes}}\vspace{-1.5em}
	\label{fig:semantic-nuscenes-visualization}

\end{figure*}

We report additional qualitative results using SalsaNext on SemKITTI, SemPOSS, and nuScenes in Figs.~\ref{fig:semantic-semkitti-visualization}, \ref{fig:semantic-semposs-visualization} and \ref{fig:semantic-nuscenes-visualization}, respectively, for both 0.1\% and 0.01\% settings.
Overall, our method surpasses SalsaNext, especially in ambiguous and cluttered regions, illustrated in Fig.~\ref{fig:semantic-semkitti-visualization}~(sidewalk/parking - row 1; vegetation/building - row 3), in Fig.~\ref{fig:semantic-semposs-visualization}~(car - row 1; rider - row 3; pole/plants - row 4), and in~Fig.~\ref{fig:semantic-nuscenes-visualization}~(terrain/other flat - row 1). Also, SalsaNext makes more mistake regarding classes with close semantical meaning as in Fig.~\ref{fig:semantic-semkitti-visualization}~(truck/car - row 1), in Fig.~\ref{fig:semantic-semposs-visualization}~(building/fence - row 5), and in Fig.~\ref{fig:semantic-nuscenes-visualization}~(bus/trailer/truck - row 2, 3, 4, 5). Furthermore, our method shows superiority in predicting small objects with similar structure or spatial position, demonstrated in Fig.~\ref{fig:semantic-semkitti-visualization} (fence/vegetation - row 4, 5; trunk/traffic sign - row 5), in Fig.~\ref{fig:semantic-semposs-visualization}~(rider/people - row 2, 3, 5; pole/plants - row 4), and in Fig.~\ref{fig:semantic-nuscenes-visualization}~(terrain/other flat - row 1). Additionally, our method infers better far away, low density regions e.g. Fig.~\ref{fig:semantic-semposs-visualization}~(car - row 1), and Fig.~\ref{fig:semantic-nuscenes-visualization}~(vegetation/barrier - row 2).

\clearpage
\newpage

\bibliography{egbib}
\end{document}